\documentclass[11pt,a4paper]{article}
\usepackage[hyperref]{naaclhlt2019}
\usepackage{times}
\usepackage{latexsym}
\usepackage{url}
\usepackage{graphicx}
\usepackage{enumitem}
\usepackage{amsmath}
\usepackage{amssymb}
\usepackage{booktabs}
\usepackage{multirow}
\usepackage{footmisc}
\usepackage{subcaption}
\usepackage{todonotes}
\usepackage{xcolor}

\DeclareMathOperator{\mean}{mean}
\DeclareMathOperator{\std}{std}

\aclfinalcopy

\title{Question Answering by Reasoning Across Documents \\ with Graph Convolutional Networks}

\author{Nicola De Cao \\
  University of Edinburgh \\
  University of Amsterdam \\
  {\tt nicola.decao@gmail.com} \\\And
  Wilker Aziz \\
  University of Amsterdam \\
  {\tt w.aziz@uva.nl} \\\And
  Ivan Titov \\
  University of Edinburgh \\
  University of Amsterdam \\
  {\tt ititov@inf.ed.ac.uk} \\}

\date{\today}

\begin{document}
\maketitle
\begin{abstract}
Most research in reading comprehension has focused on answering questions based on individual documents or even single paragraphs. We introduce a neural model which integrates and reasons relying on information spread within documents and across multiple documents. We frame it as an inference problem on a graph. Mentions of entities are nodes of this graph while edges encode relations between different mentions (e.g., within- and cross-document coreference). Graph convolutional networks (GCNs) are applied to these graphs and trained to perform multi-step reasoning. Our Entity-GCN method is scalable and compact, and it achieves state-of-the-art results on a multi-document question answering dataset, \textsc{WikiHop}~\citep{welbl2017constructing}.
\end{abstract}

\section{Introduction}

The long-standing goal of natural language understanding is the development of systems which can acquire knowledge from text collections. Fresh interest in reading comprehension tasks was sparked by the availability of large-scale datasets, such as SQuAD~\cite{rajpurkar2016squad} and CNN/Daily Mail~\cite{hermann2015teaching}, enabling end-to-end training of neural models~\cite{seo2016bidirectional,xiong2016dynamic,shen2017reasonet}. These systems, given a text and a question, need to answer the query relying on the given document. 
Recently, it has been observed that most questions in these datasets do not require reasoning across the document, but they can be answered relying on information contained in a single sentence~\cite{dirk2017fastqa}. The last generation of large-scale reading comprehension datasets, such as a NarrativeQA~\cite{kovcisky2017narrativeqa}, TriviaQA~\cite{joshi2017triviaqa}, and RACE~\cite{lai2017race}, have been created in such a way as to address this shortcoming and to ensure that systems relying only on local information cannot achieve competitive performance.

\begin{figure}[t]
    \centering
    \includegraphics[width=0.45\textwidth]{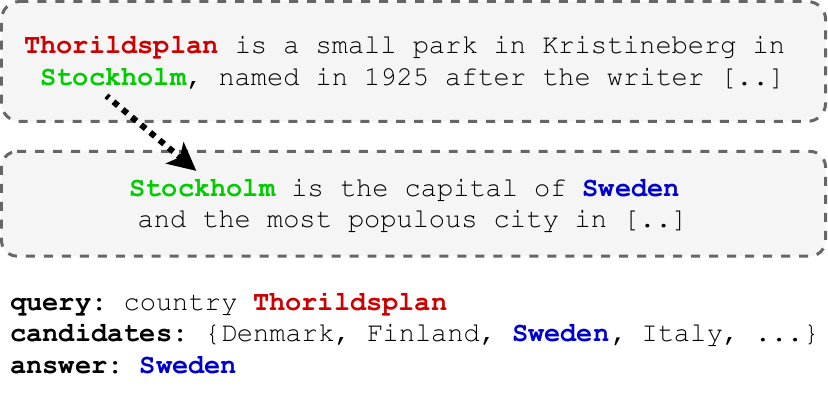}
    \caption{A sample from \textsc{WikiHop} where multi-step reasoning and information combination from different documents is necessary to infer the correct answer.}
    \label{fig:dataset}
\end{figure}

Even though these new datasets are challenging and require reasoning within documents, many question answering and search applications require aggregation of information across multiple documents. The \textsc{WikiHop} dataset~\citep{welbl2017constructing} was explicitly created to facilitate the development of systems dealing with these scenarios. Each example in \textsc{WikiHop} consists of a collection of documents, a query and a set of candidate answers (Figure~\ref{fig:dataset}). Though there is no guarantee that a question cannot be answered by relying just on a single sentence, the authors ensure that it is answerable using a chain of reasoning crossing document boundaries.

Though an important practical problem, the multi-hop setting  has so far received little attention. The methods reported by~\citet{welbl2017constructing} approach the task by merely concatenating all documents into a single long text and training a standard RNN-based reading comprehension model, namely, BiDAF~\cite{seo2016bidirectional} and FastQA~\cite{dirk2017fastqa}. Document concatenation in this setting is also used in Weaver~\cite{raison2018weaver} and MHPGM~\citep{bauer2018commonsense}. The only published paper which goes beyond concatenation is due to \citet{dhingra2018neural}, where they augment RNNs with jump-links corresponding to co-reference edges. Though these edges provide a structural bias, the RNN states are still tasked with passing the information across the document and performing multi-hop reasoning. 

Instead, we frame question answering as an inference problem on a graph representing the document collection. Nodes in this graph correspond to named entities in a document whereas edges encode relations between them (e.g., cross- and within-document coreference links or simply co-occurrence in a document). We assume that reasoning chains can be captured by propagating local contextual information along edges in this graph using a graph convolutional network (GCN)~\citep{kipf2016semi}. 

The multi-document setting imposes scalability challenges. In realistic scenarios, a system needs to learn to answer a query for a given collection (e.g., Wikipedia or a domain-specific set of documents). In such scenarios one cannot afford to run expensive document encoders (e.g., RNN or transformer-like self-attention~\cite{vaswani2017attention}), unless the computation can be preprocessed both at train and test time. Even if (similarly to  \textsc{WikiHop}  creators) one considers a coarse-to-fine approach, where a set of potentially relevant documents is provided, re-encoding them in a query-specific way remains the bottleneck. In contrast to other proposed methods (e.g., \cite{dhingra2018neural,raison2018weaver,seo2016bidirectional}), we avoid training expensive document encoders.

In our approach, only a small query encoder, the GCN layers and a simple feed-forward answer selection component are learned. Instead of training RNN encoders, we use contextualized embeddings (ELMo) to obtain initial (local) representations of nodes. This implies that only a lightweight computation has to be performed online, both at train and test time, whereas the rest is preprocessed. Even in the somewhat contrived \textsc{WikiHop} setting, where fairly small sets of candidates are provided, the model is at least 5 times faster to train than BiDAF.\footnote{When compared to the `small' and hence fast BiDAF model reported in~\citet{welbl2017constructing}, which is 25\% less accurate than our Entity-GCN. Larger RNN models are problematic also because of GPU memory constraints.} Interestingly, when we substitute ELMo with simple pre-trained word embeddings, Entity-GCN still performs on par with many techniques that use expensive question-aware recurrent document encoders.

Despite not using recurrent document encoders,
the full Entity-GCN model achieves over 2\% improvement over the best previously-published results. As our model is efficient, we also reported results of an ensemble which brings further 3.6\% of improvement and only 3\% below the human performance reported by~\citet{welbl2017constructing}. Our contributions can be summarized as follows:
\begin{itemize}
    \item we present a novel approach for multi-hop QA that relies on a (pre-trained) document encoder and information propagation across multiple documents using graph neural networks;
    \item we provide an efficient training technique which relies on a slower offline and a faster on-line computation that does not require expensive document processing;
    \item we empirically show that our algorithm is effective, presenting an improvement over previous results.
\end{itemize}

\section{Method}
In this section we explain our method. We first introduce the dataset we focus on, \textsc{WikiHop} by~\citet{welbl2017constructing}, as well as the task abstraction. We then present the building blocks that make up our Entity-GCN model, namely, an \emph{entity graph} used to relate mentions to entities within and across documents, a \emph{document encoder} used to obtain representations of mentions in context, and a \emph{relational graph convolutional network} that propagates information through the entity graph. 

\subsection{Dataset and Task Abstraction} \label{sec:data_task}
\paragraph{Data}
The \textsc{WikiHop} dataset comprises of tuples $\langle q, S_q, C_q, a^\star \rangle$ where: $q$ is a query/question, $S_q$ is a set of supporting documents, $C_q$ is a set of candidate answers (all of which are entities mentioned in $S_q$), and $a^\star \in C_q$ is the entity that correctly answers the question. \textsc{WikiHop} is assembled assuming that there exists a corpus and a knowledge base (KB) related to each other. The KB contains triples $\langle s, r, o \rangle$ where $s$ is a subject entity, $o$ an object entity, and $r$ a unidirectional relation between them. \citet{welbl2017constructing} used \textsc{Wikipedia} as corpus and \textsc{Wikidata}~\citep{vrandevcic2012wikidata} as KB. The KB is only used for constructing \textsc{WikiHop}: \citet{welbl2017constructing} retrieved the supporting documents $S_q$ from the corpus looking at mentions of subject and object entities in the text. Note that the set $S_q$  (not the KB) is provided to the QA system, and not all of the supporting documents are relevant for the query but some of them act as distractors. Queries, on the other hand, are not expressed in natural language, but instead consist of tuples $\langle s, r, ? \rangle$ where the object entity is unknown and it has to be inferred by reading the support documents. Therefore, answering a query corresponds to finding the entity $a^\star$ that is the object of a tuple in the KB with subject $s$ and relation $r$ among the provided set of candidate answers $C_q$.

\paragraph{Task}
The goal is to learn a model that can identify the correct answer $a^\star$ from the set of supporting documents $S_q$. To that end, we exploit the available supervision to train a neural network that computes scores for candidates in $C_q$. We estimate the parameters of the architecture by maximizing the likelihood of observations. For prediction, we then output the candidate that achieves the highest probability. In the following, we present our model discussing the design decisions that enable multi-step reasoning and an efficient computation.

\subsection{Reasoning on an Entity Graph} \label{sec:method_overview}
\paragraph{Entity graph}
In an offline step, we organize the content of each training instance in a graph connecting mentions of candidate answers within and across supporting documents. For a given query $q = \langle s, r, ? \rangle$, we identify mentions in $S_q$ of the entities in $C_q \cup \{s\}$ and create one node per mention. This process is based on the following heuristic:
\begin{enumerate}
    \item we consider mentions spans in $S_q$ exactly matching an element of $C_q \cup \{s\}$. Admittedly, this is a rather simple strategy which may suffer from low recall.
    \item we use predictions from a coreference resolution system to add mentions of elements in $C_q \cup \{s\}$ beyond exact matching (including both noun phrases and anaphoric pronouns). In particular, we use the end-to-end coreference resolution by~\citet{kenton2017e2e}.
    \item we discard mentions which are ambiguously resolved to multiple coreference chains; this may sacrifice recall, but avoids propagating ambiguity.
\end{enumerate}

\begin{figure}[t]
    \centering
    \includegraphics[width=0.45\textwidth]{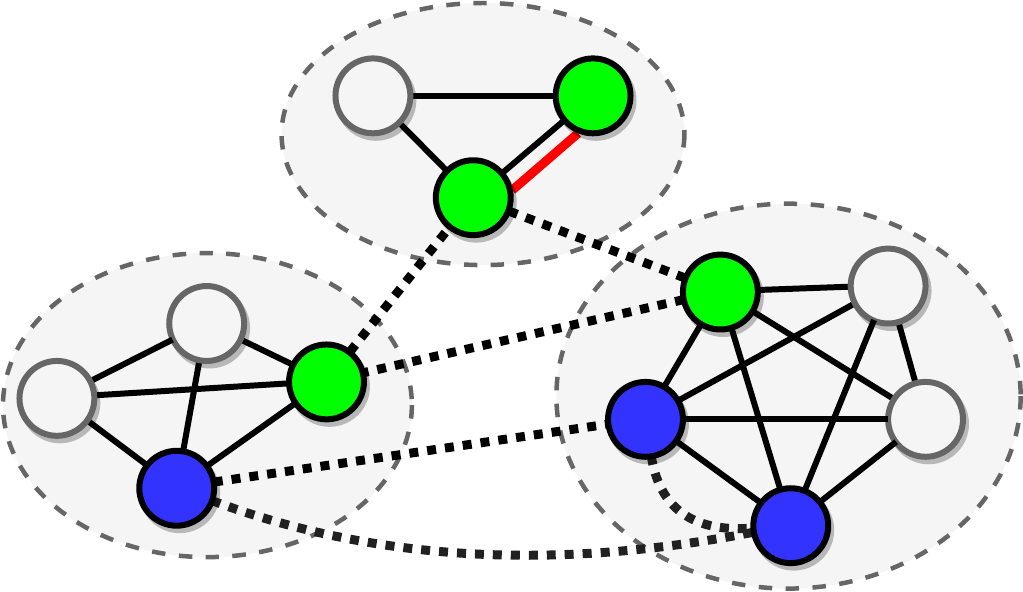}
    \caption{Supporting documents (dashed ellipses) organized as a graph where nodes are mentions of either candidate entities or query entities.  Nodes with the same color indicates they refer to the same entity (exact match, coreference or both). Nodes are connected by three simple relations: one indicating co-occurrence in the same document (solid edges), another connecting mentions that exactly match (dashed edges), and a third one indicating a coreference (bold-red line).}
    \label{fig:graph}
\end{figure}

To each node $v_i$, we associate a continuous annotation $\mathbf x_i \in \mathbb R^D$ which represents an entity in the context where it was mentioned (details in Section~\ref{sec:elmo}). We then proceed to connect these mentions i) if they co-occur within the same document (we will refer to this as \text{\texttt{DOC-BASED}} edges), ii) if the pair of named entity mentions is identical (\texttt{MATCH} edges---these may connect nodes across and within documents), or iii) if they are in the same coreference chain, as predicted by the external coreference system (\texttt{COREF}  edges). Note that \texttt{MATCH} edges when connecting mentions in the same document are mostly included in the set of edges predicted by the coreference system. Having the two types of edges lets us distinguish between less reliable edges provided by the coreference system and more reliable (but also more sparse) edges given by the exact-match heuristic. We treat these three types of connections as three different types of relations. See Figure~\ref{fig:graph} for an illustration. In addition to that, and to prevent having disconnected graphs, we add a fourth type of relation (\texttt{COMPLEMENT} edge) between any two nodes that are not connected with any of the other relations. We can think of these edges as those in the complement set of the entity graph with respect to a fully connected graph.

\paragraph{Multi-step reasoning}
Our model then approaches multi-step reasoning by transforming node representations (Section~\ref{sec:elmo} for details) with a differentiable message passing algorithm that propagates information through the entity graph. The algorithm is parameterized by a graph convolutional network (GCN)~\citep{kipf2016semi}, in particular, we employ relational-GCNs~\citep{schlichtkrull2017modeling}, an extended version that accommodates edges of different types. In Section~\ref{sec:gcn} we describe the propagation rule.

Each step of the algorithm (also referred to as a \textit{hop}) updates all node representations in parallel. In particular, a node is updated as a function of messages from its direct neighbours, and a message is possibly specific to a certain relation. At the end of the first step, every node is aware of every other node it connects directly to. Besides, the neighbourhood of a node may include mentions of the same entity as well as others (e.g., same-document relation), and these mentions may have occurred in different documents. Taking this idea recursively, each further step of the algorithm allows a node to indirectly interact with nodes already known to their neighbours. After $L$ layers of R-GCN, information has been propagated through paths connecting up to $L+1$ nodes. 

We start with node representations $\{\mathbf h_i^{(0)}\}_{i=1}^N$, and transform them by applying $L$ layers of R-GCN obtaining $\{\mathbf h_i^{(L)}\}_{i=1}^N$. Together with a representation $\mathbf q$ of the query, we define a distribution over candidate answers and we train  maximizing the likelihood of observations.
The probability of selecting a candidate $c \in C_q$ as an answer is then
\begin{equation} \label{eq:probability}
    P(c|q, C_q, S_q) \propto  \exp\left(\max_{i \in \mathcal{M}_c} f_o([\mathbf q, \mathbf h^{(L)}_i]) \right)\;, 
\end{equation}
where $f_o$ is a parameterized affine transformation, and $\mathcal{M}_c$ is the set of node indices such that $i\in \mathcal{M}_c$ only if node $v_i$ is a mention of $c$. The $\max$ operator in Equation~\ref{eq:probability} is necessary to select the node with highest predicted probability since a candidate answer is realized in multiple locations via different nodes.

\subsection{Node Annotations} \label{sec:elmo}
Keeping in mind we want an efficient model, we encode words in supporting documents and in the query using only a pre-trained model for contextualized word representations rather than training our own encoder. Specifically, we use ELMo\footnote{The use of ELMo is an implementation choice, and, in principle, any other contextual pre-trained model could be used~\cite{radford2018openai, devlin2018bert}.}~\citep{peters2018deep}, a pre-trained bi-directional language model that relies on character-based input representation. ELMo representations, differently from other pre-trained word-based models (e.g., \emph{word2vec}~\citep{mikolov2013distributed} or GloVe~\citep{pennington2014glove}), are contextualized since each token representation depends on the entire text excerpt (i.e., the whole sentence). 

We choose not to fine tune nor propagate gradients through the ELMo architecture, as it would have defied the goal of not having specialized RNN encoders. In the experiments, we will also ablate the use of ELMo showing how our model behaves using non-contextualized word representations (we use GloVe).

\paragraph{Documents pre-processing}
ELMo encodings are used to produce a set of representations $\{\mathbf x_i\}_{i=1}^N$, where $\mathbf x_i \in \mathbb R^D$ denotes the $i$th candidate mention in context. Note that these representations do not depend on the query yet and no trainable model was used to process the documents so far, that is, we use ELMo as a fixed pre-trained encoder. Therefore, we can pre-compute representation of mentions once and store them for later use.

\paragraph{Query-dependent mention encodings}
ELMo encodings are used to produce a query representation $\mathbf q \in \mathbb R^K$ as well. Here, $\mathbf q$ is a concatenation of the final outputs from a bidirectional RNN layer trained to re-encode ELMo representations of words in the query. The vector $\mathbf{q}$ is used to compute a query-dependent representation of mentions $\{\mathbf{ \hat x}_i\}_{i=1}^N$ as well as to compute a probability distribution over candidates (as in Equation~\ref{eq:probability}). Query-dependent mention encodings $\mathbf{\hat x}_i = f_x(\mathbf{q}, \mathbf{x}_i)$ are generated by a trainable function $f_x$ which is parameterized by a feed-forward neural network.

\subsection{Entity Relational Graph Convolutional Network} \label{sec:gcn}
Our model uses a gated version of the original R-GCN propagation rule. At the first layer, all hidden node representation are initialized with the query-aware encodings $\mathbf{h}_i^{(0)} = \mathbf{\hat x}_i$. Then, at each layer $0\leq \ell \leq L$, the update message $\mathbf u_i^{(\ell)}$ to the $i$th node is a sum of a transformation $f_s$ of the current node representation $\mathbf h^{(\ell)}_i$ and transformations of its neighbours:
\begin{equation}
\mathbf{u}^{(\ell)}_i = f_s(\mathbf{h}^{(\ell)}_i) + \frac{1}{|\mathcal{N}_i|} \sum_{j \in \mathcal{N}_i}  \sum_{r \in \mathcal{R}_{ij}} f_r(\mathbf{h}_j^{(\ell)})\;,
\end{equation}
where $\mathcal{N}_i$ is the set of indices of nodes neighbouring the $i$th node, $\mathcal{R}_{ij}$ is the set of edge annotations between $i$ and $j$, and $f_r$ is a parametrized function specific to an edge type $r\in\mathcal{R}$. Recall the available relations from  Section~\ref{sec:method_overview}, namely, $\mathcal{R} =\{$\text{\texttt{DOC-BASED}}, \texttt{MATCH}, \texttt{COREF}, \texttt{COMPLEMENT}$\}$. 

A gating mechanism regulates how much of the update message propagates to the next step. This provides the model a way to prevent completely overwriting past information. Indeed, if all necessary information to answer a question is present at a layer which is not the last, then the model should learn to stop using neighbouring information for the next steps. Gate levels are computed as
\begin{equation}
\mathbf{a}^{(\ell)}_i = \sigma \left( f_a\left([\mathbf{u}^{(\ell)}_i, \mathbf{h}^{(\ell)}_i ]\right) \right) \;,
\end{equation}
where $\sigma(\cdot)$ is the sigmoid function and $f_a$ a parametrized transformation. Ultimately, the updated representation is a gated combination of the previous representation and a non-linear transformation of the update message:
\begin{equation}
\mathbf{h}^{(\ell + 1)}_i = \phi(\mathbf{u}^{(\ell)}_i) \odot \mathbf{a}^{(\ell)}_i + \mathbf{h}^{(\ell)}_i \odot (1 - \mathbf{a}^{(\ell)}_i ) \;,
\end{equation}
where $\phi(\cdot)$ is any nonlinear function (we used $\tanh$) and $\odot$ stands for element-wise multiplication. All transformations $f_*$ are affine and they are not layer-dependent (since we would like to use as few parameters as possible to decrease model complexity promoting efficiency and scalability).

\section{Experiments}
In this section, we compare our method against recent work as well as preforming an ablation study using the \textsc{WikiHop} dataset~\citep{welbl2017constructing}. See Appendix~\ref{app:architecture} in the supplementary material for a description of the hyper-parameters of our model and training details.

\paragraph{\textsc{WikiHop}}
We use \textsc{WikiHop} for training, validation/development and test. The test set is not publicly available and therefore we measure performance on the validation set in almost all experiments. \textsc{WikiHop} has 43,738/ 5,129/ 2,451 query-documents samples in the training, validation and test sets respectively for a total of 51,318 samples. Authors constructed the dataset as described in Section~\ref{sec:data_task} selecting samples with a graph traversal up to a maximum chain length of 3 documents (see Table~\ref{tab:dataset} for additional dataset statistics). \textsc{WikiHop} comes in two versions, a standard (unmasked) one and a masked one. The masked version was created by the authors to test whether methods are able to learn lexical abstraction. In this version, all candidates and all mentions of them in the support documents are replaced by random but consistent placeholder tokens. Thus, in the masked version, mentions are always referred to via unambiguous surface forms. We do not use coreference systems in the masked version as they rely crucially on lexical realization of mentions and cannot operate on masked tokens.

\begin{table}
    \centering
    \begin{tabular}{lrrrr}
    \toprule
    & \textbf{Min} & \textbf{Max} & \textbf{Avg.}  &  \textbf{Median} \\
    \midrule
    \# candidates & 2 & 79 & 19.8 & 14 \\
    \# documents & 3 & 63 & 13.7 & 11  \\
    \# tokens/doc. & 4 & 2,046 & 100.4 & 91  \\
    \bottomrule
    \end{tabular}
    \caption{\textsc{WikiHop} dataset statistics from \citet{welbl2017constructing}: number of candidates and documents per sample and document length.}
    \label{tab:dataset}
\end{table}

\subsection{Comparison} \label{sec:comparison}
In this experiment, we compare our Enitity-GCN against recent prior work on the same task. We present test and development results (when present) for both versions of the dataset in Table~\ref{tab:comparison}. From~\citet{welbl2017constructing}, we list an oracle based on human performance as well as two standard reading comprehension models, namely BiDAF~\citep{seo2016bidirectional} and FastQA~\citep{dirk2017fastqa}. We also compare against Coref-GRU~\citep{dhingra2018neural}, MHPGM~\citep{bauer2018commonsense}, and Weaver~\citep{raison2018weaver}. Additionally, we  include results of MHQA-GRN~\cite{song2018exploring}, from a recent arXiv preprint describing concurrent work. They jointly train graph neural networks and recurrent encoders. We report single runs of our two best single models and an ensemble one on the unmasked test set (recall that the test set is not publicly available and the task organizers only report unmasked results) as well as both versions of the validation set.

\begin{table*}[t]
\centering
\begin{tabular}{l|cc|cc}
\toprule
\multirow{2}{*}{\textbf{Model}} & \multicolumn{2}{c}{Unmasked} & \multicolumn{2}{c}{Masked} \\
 & Test & Dev & Test & Dev \\
\midrule
Human~\citep{welbl2017constructing} & 74.1 & -- & -- & -- \\
FastQA~\citep{welbl2017constructing} & 25.7 & -- & 35.8 & -- \\
BiDAF~\citep{welbl2017constructing} & 42.9 & -- & 54.5 & -- \\
Coref-GRU~\citep{dhingra2018neural} & 59.3 & 56.0 & -- & -- \\
MHPGM~\citep{bauer2018commonsense} & -- & 58.2 & -- & -- \\
Weaver / Jenga~\citep{raison2018weaver} & 65.3 & 64.1  & -- & -- \\
MHQA-GRN~\citep{song2018exploring} & 65.4 & 62.8  & -- & -- \\
\midrule
Entity-GCN without coreference (single model) & \textbf{67.6} & \textbf{64.8} & -- & 70.5 \\
Entity-GCN with coreference (single model) & \textbf{66.4} &\textbf{65.3} & -- & -- \\
Entity-GCN* (ensemble 5 models) & \textbf{71.2} & \textbf{68.5} & -- & 71.6 \\
\midrule
\end{tabular}
\caption{Accuracy of different models on \textsc{WikiHop} closed test set and public validation set. Our Entity-GCN outperforms recent prior work without learning any language model to process the input but relying on a pre-trained one (ELMo -- without fine-tunning it) and applying R-GCN to reason among entities in the text. *  with coreference for unmasked dataset and without coreference for the masked one.}
\label{tab:comparison}
\end{table*}

Entity-GCN (best single model without coreference edges) outperforms all previous work by over 2\% points. We additionally re-ran BiDAF baseline to compare training time: when using a single Titan X GPU, BiDAF and Entity-GCN process 12.5 and 57.8 document sets per second, respectively. Note that~\citet{welbl2017constructing} had to use BiDAF with very small state dimensionalities (20), and smaller batch size due to the scalability issues (both memory and computation costs). We compare applying the same reductions.\footnote{Besides, we could not run any other method we compare with combined with ELMo without reducing the dimensionality further or having to implement a distributed version.} Eventually, we also report an ensemble of 5 independently trained models. All models are trained on the same dataset splits with different weight initializations. The ensemble prediction is obtained as $\arg\max\limits_c \prod\limits_{i=1}^5 P_i(c|q, C_q, S_q)$ from each model.

\subsection{Ablation Study} \label{sec:ablation}
To help determine the sources of improvements, we perform an ablation study using the publicly available validation set (see Table~\ref{tab:ablation}). We perform two groups of ablation, one on the embedding layer, to study the effect of ELMo, and one on the edges, to study how different relations affect the overall model performance.

\paragraph{Embedding ablation}
We argue that ELMo is crucial, since we do not rely on any other context encoder. However, it is interesting to explore how our R-GCN performs without it.
Therefore, in this experiment, we replace the deep contextualized embeddings of both the query and the nodes with GloVe~\citep{pennington2014glove} vectors (insensitive to context). Since we do not have any component in our model that processes the documents, we expect a drop in performance. In other words, in this ablation our model tries to answer questions \emph{without reading the context at all}. For example, in Figure~\ref{fig:dataset}, our model would be aware that ``Stockholm'' and ``Sweden'' appear in the same document but any context words, including the ones encoding relations (e.g., ``is the capital of'') will be hidden. Besides, in the masked case all mentions become `unknown' tokens with GloVe and therefore the predictions are equivalent to a random guess. Once the strong pre-trained encoder is out of the way, we also ablate the use of our R-GCN component, thus completely depriving the model from inductive biases that aim at multi-hop reasoning.

The first important observation is that replacing ELMo by GloVe (GloVe with R-GCN in Table~\ref{tab:ablation}) still yields a competitive system that ranks far above baselines from~\citep{welbl2017constructing} and even above the Coref-GRU of~\citet{dhingra2018neural}, in terms of accuracy on (unmasked) validation set. The second important observation is that if we then remove R-GCN (GloVe w/o R-GCN in Table~\ref{tab:ablation}), we lose 8.0 points. That is, the R-GCN component pushes the model to perform above Coref-GRU still without accessing context, but rather by updating mention representations based on their relation to other ones. These results highlight the impact of our R-GCN component.

\paragraph{Graph edges ablation}
In this experiment we investigate the effect of the  different relations available in the entity graph and processed by the R-GCN module. We start off by testing our stronger encoder (i.e., ELMo) in absence of edges connecting mentions in the supporting documents (i.e., using only self-loops -- No R-GCN in Table~\ref{tab:ablation}). The results suggest that \textsc{WikipHop} genuinely requires multihop inference, as our best model is 6.1\% and 8.4\% more accurate than this local model, in unmasked and masked settings, respectively.\footnote{Recall that all models in the ensemble use the same local representations, ELMo.} However, it also shows that ELMo representations capture predictive context features, without being explicitly trained for the task. It confirms that our goal of getting away with training expensive document encoders is a realistic one.

\begin{table}[t]
\centering
\begin{tabular}{l|ccc}
\toprule
\textbf{Model} & unmasked & masked \\
\midrule
\emph{full} (ensemble) & \textbf{68.5} & \textbf{71.6} \\ 
\emph{full} (single) & \textbf{65.1 {\scriptsize $\pm$ 0.11}} & \textbf{70.4 {\scriptsize $\pm$ 0.12}} \\
\midrule
GloVe with R-GCN & 59.2 & 11.1 \\
GloVe w/o R-GCN & 51.2 & 11.6 \\
\midrule
No R-GCN  & 62.4 & 63.2 \\
No relation types  & 62.7 & 63.9 \\
No \texttt{DOC-BASED}  & 62.9 & 65.8 \\ 
No \texttt{MATCH}  & 64.3 & 67.4 \\ 
No \texttt{COREF}  & 64.8 & -- \\
No \texttt{COMPLEMENT} & 64.1 & 70.3 \\
Induced edges & 61.5 & 56.4 \\
\bottomrule
\end{tabular}
\caption{Ablation study on \textsc{WikiHop} validation set. The \emph{full model} is our Entity-GCN with all of its components and other rows indicate models trained without a component of interest. We also report baselines using GloVe instead of ELMo with and without R-GCN. For the \emph{full model} we report $\mean \pm 1 \std$ over 5 runs.}
\label{tab:ablation}
\end{table}

\begin{table*}[t]
\centering
\begin{tabular}{rl|ccccr}
\toprule
& \textbf{Relation} & \textbf{Accuracy} & \textbf{P@2} & \textbf{P@5} & \textbf{Avg. $|C_q|$} & \textbf{Supports} \\
\midrule
& overall (ensemble) & 68.5 & 81.0 & 94.1 & 20.4 {\scriptsize $\pm$ 16.6} & 5129 \\
& overall (single model) & 65.3 & 79.7 & 92.9 & 20.4 {\scriptsize $\pm$ 16.6} & 5129 \\
\midrule
\multirow{3}{*}{\textbf{3 best}} & member\_of\_political\_party & 85.5 & 95.7 & 98.6 & 5.4 {\scriptsize $\pm$ 2.4} & 70 \\
& record\_label & 83.0 & 93.6 & 99.3 & 12.4 {\scriptsize $\pm$ 6.1} & 283 \\
& publisher & 81.5 & 96.3 & 100.0 & 9.6 {\scriptsize $\pm$ 5.1} & 54 \\
\midrule
\multirow{3}{*}{\textbf{3 worst}} & place\_of\_birth & 51.0 & 67.2 & 86.8 & 27.2 {\scriptsize $\pm$ 14.5} & 309 \\
& place\_of\_death & 50.0 & 67.3 & 89.1 & 25.1 {\scriptsize $\pm$ 14.3} & 159 \\
& inception & 29.9 & 53.2 & 83.1 & 21.9 {\scriptsize $\pm$ 11.0} & 77 \\
\bottomrule
\end{tabular}
\caption{Accuracy and precision at K (P@K in the table) analysis overall and per query type. Avg. $|C_q|$ indicates the average number of candidates with one standard deviation.}
\label{tab:questions}
\end{table*}

We then inspect our model's effectiveness in making use of the structure encoded in the graph. We start naively by fully-connecting all nodes within and across documents without distinguishing edges by type (No relation types in Table~\ref{tab:ablation}). We observe only marginal improvements with respect to ELMo alone (No R-GCN in Table~\ref{tab:ablation}) in both the unmasked and masked setting suggesting that a GCN operating over a naive entity graph would not add much to this task and a more informative graph construction and/or a more sophisticated parameterization is indeed needed.

Next, we ablate each type of relations independently, that is, we either remove connections of mentions that co-occur in the same document (\texttt{DOC-BASED}), connections between mentions matching exactly (\texttt{MATCH}), or edges predicted by the coreference system (\texttt{COREF}). The first thing to note is that the model makes better use of \texttt{DOC-BASED} connections than \texttt{MATCH} or \texttt{COREF} connections. This is mostly because i) the majority of the connections are indeed between mentions in the same document, and ii) without connecting mentions within the same document we remove important information since the model is unaware they appear closely in the document. Secondly, we notice that coreference links and complement edges seem to play a more marginal role. Though it may be surprising for coreference edges, recall that the \texttt{MATCH} heuristic already captures the easiest coreference cases, and for the rest the out-of-domain coreference system may not be reliable. Still, modelling all these different relations together gives our Entity-GCN a clear advantage. This is our best system evaluating on the development. Since Entity-GCN seems to gain little advantage using the coreference system, we report test results both with and without using it. Surprisingly, with coreference, we observe performance degradation on the test set. It is likely that the test documents are harder for the coreference system.\footnote{Since the test set is hidden from us, we cannot analyze this difference further.}

We do perform one last ablation, namely, we replace our heuristic for assigning edges and their labels by a model component that predicts them. The last row of Table~\ref{tab:ablation} (Induced edges) shows model performance when edges are not predetermined but predicted. For this experiment, we use a bilinear function $f_e(\mathbf{\hat x}_i, \mathbf{\hat x}_j) = \sigma\left( \mathbf{\hat x}^\top_i \mathbf{W}_e \mathbf{\hat x}_j \right)$ that predicts the importance of a single edge connecting two nodes $i,j$ using the query-dependent representation of mentions (see Section~\ref{sec:elmo}). The performance drops below `No R-GCN' suggesting that it cannot learn these dependencies on its own.

Most results are stronger for the masked settings even though we do not apply the coreference resolution system in this setting due to masking. It is not surprising as coreferred mentions are labeled with the same identifier in the masked version, even if their original surface forms did not match (\citet{welbl2017constructing} used \textsc{Wikipedia} links for masking). Indeed, in the masked version, an entity is always referred to via the same unique surface form (e.g., \texttt{MASK1}) within and across documents. In the unmasked setting, on the other hand, mentions to an entity may differ (e.g., ``US'' vs ``United States'') and they might not be retrieved by the coreference system we are employing, making the task harder for all models. Therefore, as we rely mostly on exact matching when constructing our graph for the masked case, we are more effective in recovering coreference links on the masked rather than unmasked version.\footnote{Though other systems do not explicitly link matching mentions, they similarly benefit from masking (e.g., masks essentially single out spans that contain candidate answers).}

\section{Error Analysis} \label{sec:error}
In this section we provide an error analysis for our best single model predictions. First of all, we look at which type of questions our model performs well or poorly. There are more than 150 query types in the validation set but we filtered the three with the best and with the worst accuracy that have at least 50 supporting documents and at least 5 candidates. We show results in Table~\ref{tab:questions}. We observe that questions regarding places (birth and death) are considered harder for Entity-GCN. We then inspect samples where our model fails while assigning highest likelihood and noticed two principal sources of failure i) a mismatch between what is written in \textsc{Wikipedia} and what is annotated in \textsc{Wikidata}, and ii) a different degree of granularity (e.g., born in ``London'' vs ``UK'' could be considered both correct by a human but not when measuring accuracy). See Table~\ref{tab:error_sample} in the supplement material for some reported samples.

Secondly, we study how the model performance degrades when the input graph is large. In particular, we observe a negative Pearson's correlation (-0.687) between accuracy and the number of candidate answers. However, the performance does not decrease steeply. The distribution of the number of candidates in the dataset peaks at 5 and has an average of approximately 20. Therefore, the model does not see many samples where there are a large number of candidate entities during training. Differently, we notice that as the number of nodes in the graph increases, the model performance drops but more gently (negative but closer to zero Pearson's correlation). This is important as document sets can be large in practical applications.
See Figure~\ref{fig:results} in the supplemental material for plots.

\section{Related Work} \label{sec:related}
In previous work, BiDAF~\citep{seo2016bidirectional}, FastQA~\citep{dirk2017fastqa}, Coref-GRU~\citep{dhingra2018neural}, MHPGM~\citep{bauer2018commonsense}, and Weaver / Jenga~\citep{raison2018weaver} have been applied to multi-document question answering. The first two mainly focus on single document QA and~\citet{welbl2017constructing} adapted both of them to work with \textsc{WikiHop}. They process each instance of the dataset by concatenating all $d \in S_q$ in a random order adding document separator tokens. They trained using the first answer mention in the concatenated document and evaluating exact match at test time. Coref-GRU, similarly to us, encodes relations between entity mentions in the document. Instead of using graph neural network layers, as we do, they augment RNNs with jump links corresponding to pairs of corefereed mentions. MHPGM uses a multi-attention mechanism in combination with external commonsense relations to perform multiple hops of reasoning. Weaver is a deep co-encoding model that uses several alternating bi-LSTMs to process the concatenated documents and the query.

Graph neural networks have been shown successful on a number of NLP tasks~\cite{marcheggiani2017encoding,bastings2017graph,zhanggraph}, including those involving document level modeling~\cite{peng2017cross}. They have also been applied in the context of asking questions about knowledge contained in a knowledge base~\cite{zhang2017variational}. In \newcite{schlichtkrull2017modeling}, GCNs are used to capture reasoning chains in a knowledge base. Our work and unpublished concurrent work by~\citet{song2018exploring} are the first to study graph neural networks in the context of multi-document QA. Besides differences in the architecture, \citet{song2018exploring} propose to train a combination of a graph recurrent network and an RNN encoder. We do not train any RNN document encoders in this work.

\section{Conclusion} \label{sec:conclusion}
We designed a graph neural network that operates over a compact graph representation of a set of documents where nodes are mentions to entities and edges signal relations such as within and cross-document coreference. The model learns to answer questions by gathering evidence from different documents via a differentiable message passing algorithm that updates node representations based on their neighbourhood. Our model outperforms published results where ablations show substantial evidence in favour of multi-step reasoning. Moreover, we make the model fast by using pre-trained (contextual) embeddings.

\section*{Acknowledgments}
We would like to thank Johannes Welbl for helping to test our system on \textsc{WikiHop}. This project is supported by SAP Innovation Center Network, ERC Starting Grant BroadSem (678254) and the Dutch Organization for Scientific Research (NWO) VIDI 639.022.518. Wilker Aziz is supported by the Dutch Organisation for Scientific Research (NWO) VICI Grant nr. 277-89-002.

\bibliographystyle{acl_natbib}
\bibliography{main}

\clearpage

\appendix

\section{Implementation and Experiments Details} \label{app:architecture}

\subsection{Architecture}

See table~\ref{tab:architecture} for an outline of Entity-GCN architectural detail. Here the computational steps

\begin{enumerate}
    \item ELMo embeddings are a concatenation of three 1024-dimensional vectors resulting in 3072-dimensional input vectors $\{\mathbf x_i\}_{i=1}^N$.
    \item For the query representation $\mathbf q$, we apply 2 bi-LSTM layers of 256 and 128 hidden units to its ELMo vectors. The concatenation of the forward and backward states results in a 256-dimensional question representation.
    \item ELMo embeddings of candidates are projected to 256-dimensional vectors, concatenated to the $\mathbf q$, and further transformed with a two layers MLP of 1024 and 512 hidden units in 512-dimensional query aware entity representations $\{\mathbf{\hat x}_i\}_{i=1}^N \in \mathbb{R}^{512}$.
    \item All transformations $f_*$ in R-GCN-layers are affine and they do maintain the input and output dimensionality of node representations the same (512-dimensional).
    \item Eventually, a 2-layers MLP with [256, 128] hidden units takes the concatenation between $\{\mathbf{h}_i^{(L)}\}_{i=1}^N$ and $\mathbf{q}$ to predict the probability that a candidate node $v_i$ may be the answer to the query $q$ (see Equation~\ref{eq:probability}).
\end{enumerate}

During preliminary trials, we experimented with different numbers of R-GCN-layers (in the range 1-7). We observed that with \textsc{WikiHop}, for $L \geq 3$ models reach essentially the same performance, but more layers increase the time required to train them. Besides, we observed that the gating mechanism learns to keep more and more information from the past at each layer making unnecessary to have more layers than required.

\begin{table*}[t]
\centering
\begin{tabular}{c|c}
\toprule
\multicolumn{2}{c}{\textbf{Input} - q, $\{v_i\}_{i=1}^N$} \\
\midrule
query ELMo 3072-dim & candidates ELMo 3072-dim\\
\midrule
2 layers bi-LSTM [256, 128]-dim & 1 layer FF 256-dim \\
\midrule
\multicolumn{2}{c}{concatenation 512-dim} \\
\midrule
\multicolumn{2}{c}{2 layer FF [1024, 512]-dim: : $\{\mathbf{\hat x}_i\}_{i=1}^N$} \\
\midrule
\multicolumn{2}{c}{3 layers R-GCN 512-dim each (shared parameters)} \\
\midrule
\multicolumn{2}{c}{concatenation with  $\mathbf{q}$ 768-dim} \\
\midrule
\multicolumn{2}{c}{3 layers FF [256,128,1]-dim} \\
\midrule
\multicolumn{2}{c}{\textbf{Output} - probabilities over $C_q$} \\
\bottomrule
\end{tabular}
\caption{Model architecture.}
\label{tab:architecture}
\end{table*}

\subsection{Training Details}

We train our models with a batch size of 32 for at most 20 epochs using the Adam optimizer~\citep{kingma2014adam} with $\beta_1=0.9$, $\beta_2=0.999$ and a learning rate of $10^{-4}$. To help against overfitting, we employ dropout (drop rate $\in {0, 0.1, 0.15, 0.2, 0.25}$) ~\citep{srivastava2014dropout} and early-stopping on validation accuracy. We report the best results of each experiment based on accuracy on validation set.

\section{Error Analysis}

In Table~\ref{tab:error_sample}, we report three samples from \textsc{WikiHop} development set where out Entity-GCN fails. In particular, we show two instances where our model presents high confidence on the answer, and one where is not. We commented these samples explaining why our model might fail in these cases.

\begin{table*}[ht]

\begin{subtable}{\textwidth}
\centering
\begin{tabular}{r|l|r|l}
\toprule
\textbf{ID} & WH\_dev\_2257 & \textbf{Gold answer} & 2003 ($p=14.1$)  \\
\midrule
\textbf{Query} & inception (of) Derrty Entertainment & \textbf{Predicted answer} & 2000 ($p=15.8$) \\
\midrule
\textbf{Support 1} & \multicolumn{3}{p{35em}}{\textbf{Derrty Entertainment} is a record label founded by [...]. The first album released under \textbf{Derrty Entertainment} was Nelly 's \textbf{Country Grammar}. } \\
\midrule
\textbf{Support 2} & \multicolumn{3}{p{35em}}{
\textbf{Country Grammar} is the debut single by American rapper Nelly. The song was produced by Jason Epperson. It was released in \textbf{2000}, [...]
} \\
\bottomrule
\end{tabular}
\caption{In this example, the model predicts the answer correctly. However, there is a mismatch between what is written in \textsc{Wikipedia} and what is annotated in \textsc{Wikidata}. In \textsc{WikiHop}, answers are generated with \textsc{Wikidata}.}
\end{subtable}

\vspace{.5em}

\begin{subtable}{\textwidth}
\centering
\begin{tabular}{r|l|r|l}
\toprule
\textbf{ID} & WH\_dev\_2401 & \textbf{Gold answer} & Adolph Zukor ($p=7.1$e$-4\%$)  \\
\midrule
\textbf{Query} & producer (of) Forbidden Paradise & \textbf{Predicted answer} & Jesse L. Lask ($p=99.9\%$) \\
\midrule
\textbf{Support 1} & \multicolumn{3}{p{35em}}{\textbf{Forbidden Paradise} is a [...] drama film produced by \textbf{Famous Players-Lasky} [...] } \\
\midrule
\textbf{Support 2} & \multicolumn{3}{p{35em}}{\textbf{Famous Players-Lasky} Corporation was [...] from the merger of \textbf{Adolph Zukor}'s Famous Players Film Company [..] and the \textbf{Jesse L. Lasky} Feature Play Company. } \\
\bottomrule
\end{tabular}
\caption{In this sample, there is ambiguity between two entities since both are correct answers reading the passages but only one is marked as correct. The model fails assigning very high probability to only on one of them.}
\end{subtable}

\vspace{.5em}

\begin{subtable}{\textwidth}
\centering
\begin{tabular}{r|l|r|l}
\toprule
\textbf{ID} & WH\_dev\_3030 & \textbf{Gold answer} &  Scania ($p= 0.029\%$) \\
\midrule
\textbf{Query} & place\_of\_birth (of) Erik Penser & \textbf{Predicted answer} & Esl\"ov ($p=97.3\%$) \\
\midrule
\textbf{Support 1} & \multicolumn{3}{p{35em}}{Nils Wilhelm \textbf{Erik Penser} (born August 22, 1942, in \textbf{Esl\"ov}, \textbf{Sk\r{a}ne}) is a Swedish [...] } \\
\midrule
\textbf{Support 2} & \multicolumn{3}{p{35em}}{\textbf{Sk\r{a}ne} County, sometimes referred to as `` \textbf{Scania} County '' in English, is the [...] } \\
\bottomrule
\end{tabular}
\caption{In this sample, there is ambiguity between two entities since the city Esl\"ov is located in the Scania County (English name of Sk\r{a}ne County). The model assigning high probability to the city and it cannot select the county.}
\end{subtable}

\caption{Samples from \textsc{WikiHop} set where Entity-GCN fails. $p$ indicates the predicted likelihood.}
\label{tab:error_sample}
\end{table*}

\section{Ablation Study}

\begin{figure}[h]
\centering
\begin{subfigure}[b]{0.48\textwidth}
    \includegraphics[width=\textwidth]{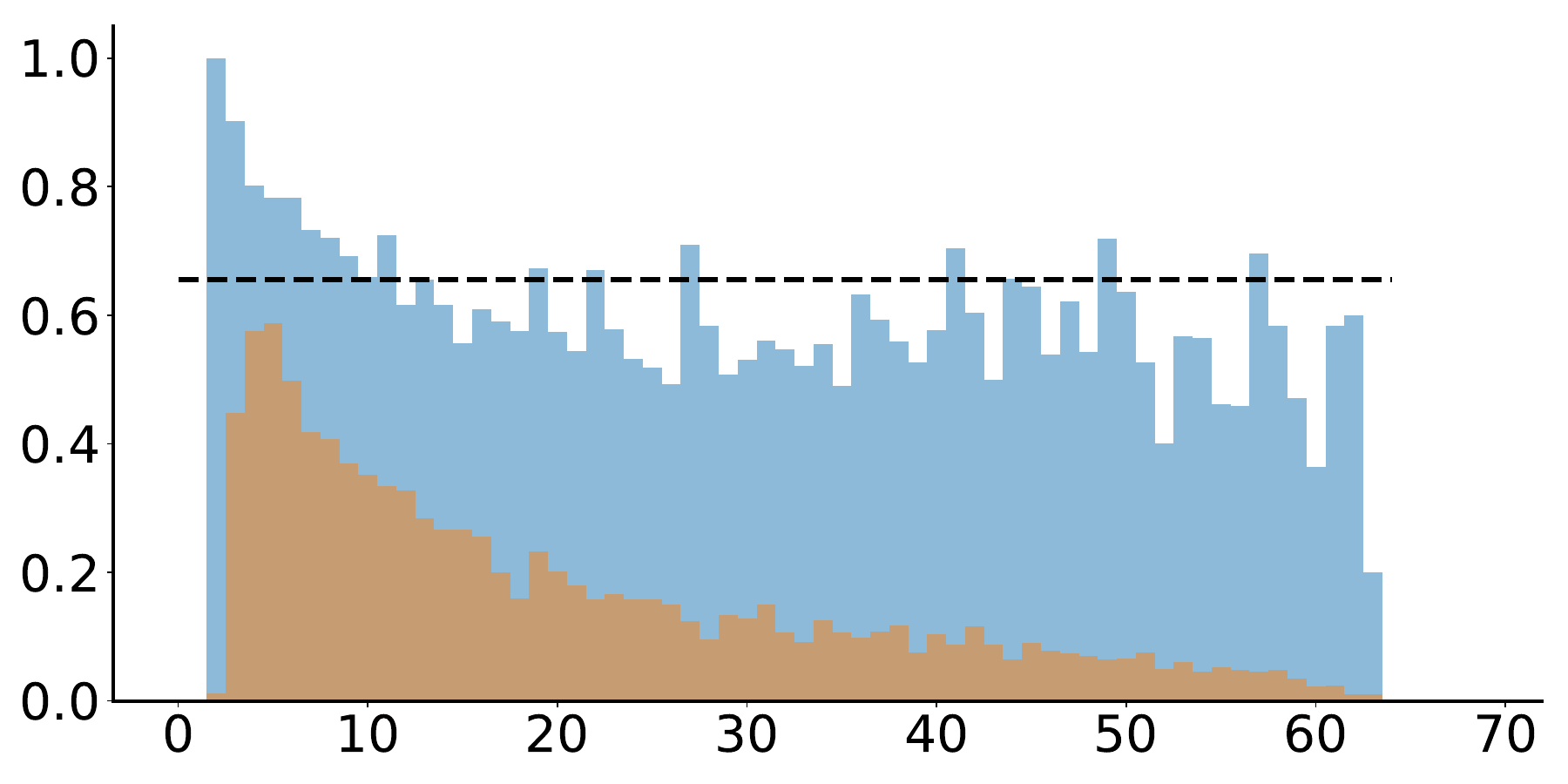}
    \caption{Candidates set size (x-axis) and accuracy (y-axis). Pearson's correlation of $-0.687$ ($p<10^{-7}$).}
    \label{fig:cadidates_len}
\end{subfigure}
~
\begin{subfigure}[b]{0.48\textwidth}
    \includegraphics[width=\textwidth]{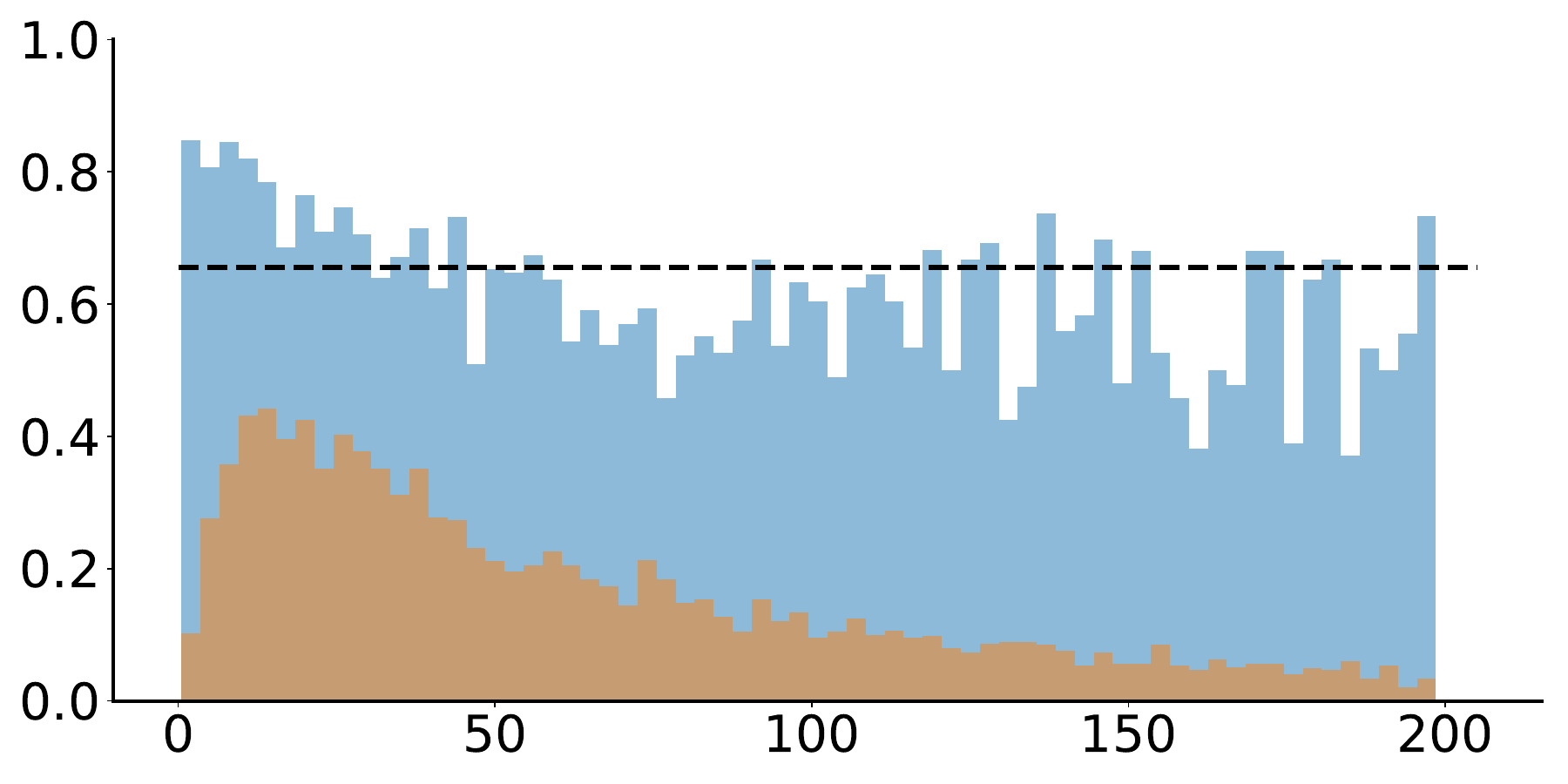}
    \caption{Nodes set size (x-axis) and accuracy (y-axis). Pearson's correlation of $-0.385$ ($p<10^{-7}$).}
    \label{fig:nodes_lens}
\end{subfigure}
\caption{Accuracy (blue) of our best single model with respect to the candidate set size (on the \emph{top}) and nodes set size (on the \emph{bottom}) on the validation set. Re-scaled data distributions (orange) per number of candidate (\emph{top}) and nodes (\emph{bottom}). Dashed lines indicate  average accuracy.}
\label{fig:results}
\end{figure}

In Figure~\ref{fig:results}, we show how the model performance goes when the input graph is large. In particular, how Entity-GCN performs as the number of candidate answers or the number of nodes increases.

\end{document}